\documentclass{article}
\pdfoutput=1

\usepackage[utf8]{inputenc}
\usepackage[nonatbib,final]{nips_2017}

\usepackage[utf8]{inputenc} % allow utf-8 input
\usepackage[T1]{fontenc}    % use 8-bit T1 fonts
\usepackage{url}            % simple URL typesetting
\usepackage{booktabs}       % professional-quality tables
\usepackage{amsfonts}       % blackboard math symbols
\usepackage{nicefrac}       % compact symbols for 1/2, etc.
\usepackage{microtype}      % microtypography

\usepackage{tgtermes}
\usepackage{amsmath,amsthm,bm,amsfonts}
\usepackage{enumerate}
\usepackage{scalefnt,letltxmacro}
\LetLtxMacro{\oldtextsc}{\textsc}
\renewcommand{\textsc}[1]{\oldtextsc{\scalefont{1.10}#1}}
\usepackage[scaled=0.92]{PTSans}
\usepackage{booktabs}
\usepackage{caption}

\usepackage[usenames,dvipsnames]{xcolor}
\definecolor{shadecolor}{gray}{0.9}

\usepackage{titlesec}
\titlespacing\section{0pt}{2pt plus 2pt minus 2pt}{2pt plus 2pt minus 2pt}
\titlespacing\subsection{0pt}{2pt plus 2pt minus 2pt}{2pt plus 2pt minus 2pt}
\titlespacing\subsubsection{0pt}{0pt plus2pt minus 2pt}{0pt plus 2pt minus 2pt}
\titlespacing\paragraph{0pt}{0pt plus 2pt minus 2pt}{0pt plus 2pt minus 2pt}

\usepackage{ragged2e}

\usepackage{graphicx}
\usepackage[labelfont=bf]{caption}
\usepackage{subfigure}

\usepackage{booktabs}
\usepackage{tabu}

\usepackage{natbib}

\usepackage[algoruled]{algorithm2e}
\usepackage{listings}
\usepackage{fancyvrb}
\fvset{fontsize=\normalsize}

\usepackage[colorlinks,linktoc=all]{hyperref}
\usepackage[all]{hypcap}
\hypersetup{citecolor=Violet}
\hypersetup{linkcolor=black}
\hypersetup{urlcolor=MidnightBlue}

\usepackage
[acronym,smallcaps,nowarn,section,nogroupskip,nonumberlist]{glossaries}
\glsdisablehyper{}

\usepackage{listings}
\lstdefinestyle{alp_style}{
    commentstyle=\color{OliveGreen},
    numberstyle=\tiny\color{black!60},
    stringstyle=\color{BrickRed},
    basicstyle=\ttfamily\scriptsize,
    breakatwhitespace=false,
    breaklines=true,
    captionpos=b,
    keepspaces=true,
    numbers=none,
    numbersep=5pt,
    showspaces=false,
    showstringspaces=false,
    showtabs=false,
    tabsize=2
}
\usepackage{tikz}
\usetikzlibrary{arrows,shadows,shapes,backgrounds,decorations,snakes,fit}
\DeclareMathOperator*{\E}{\mathbb{E}}

\DeclareMathOperator*{\bx}{{\bf x}}
\DeclareMathOperator*{\bz}{{\bf z}}

\usepackage{tikz}
\usetikzlibrary{arrows,shadows,shapes,backgrounds,decorations,snakes,fit}

\title{Perturbative Black Box Variational Inference}
\author{Robert Bamler\thanks{Equal contributions. First authorship determined by coin flip among first two authors.} \\ Disney Research \\Pittsburgh, USA\And
Cheng Zhang$^\ast$ \\ Disney Research \\Pittsburgh, USA \And
Manfred Opper \\ TU Berlin \\Berlin, Germany\And
Stephan Mandt$^\ast$ \\ Disney Research \\Pittsburgh, USA \AND
{\tt firstname.lastname$@$\{disneyresearch.com, tu-berlin.de\}}
}
\begin{document}
\maketitle

\begin{abstract}
Black box variational inference (BBVI) with reparameterization gradients triggered the exploration of divergence measures other than the Kullback-Leibler (KL) divergence, such as alpha divergences.
In this paper, we view BBVI with generalized divergences as a form of estimating the marginal likelihood via biased importance sampling.
The choice of divergence determines a bias-variance trade-off between the tightness of a bound on the marginal likelihood (low bias) and the variance of its gradient estimators.
Drawing on variational perturbation theory of statistical physics, we use these insights to construct a family of new variational bounds.
Enumerated by an odd integer order $K$, this family captures the standard KL bound for $K=1$, and converges to the exact marginal likelihood as $K\to\infty$.
Compared to alpha-divergences, our reparameterization gradients have a lower variance.
We show in experiments on Gaussian Processes and Variational Autoencoders that the new bounds are more mass covering, and that the resulting posterior covariances are closer to the true posterior and lead to higher likelihoods on held-out data.
\end{abstract}

\section{Introduction}

Variational inference (VI) \citep{jordan1999introduction} provides a way to convert Bayesian inference to optimization by minimizing a divergence measure. Recent advances of VI have been devoted to scalability~\citep{hoffman2013stochastic,ranganath2014black}, divergence measures~\citep{minka2005divergence,li2016renyi, hernandez2016black}, and structured variational distributions \citep{hoffman2015stochastic,ranganath2016hierarchical}.

While traditional stochastic variational inference (SVI) \citep{hoffman2013stochastic} was limited to conditionally conjugate Bayesian models,  black box variational inference (BBVI) \citep{ranganath2014black} enables SVI on a large class of models.
It expresses the gradient as an expectation, and estimates it by Monte-Carlo sampling.
A variant of BBVI uses reparameterized gradients and has lower variance~\citep{salimans2013fixed,kingma2013auto,rezende2014stochastic,ruiz2016generalized}.
BBVI paved the way for approximate inference in complex and deep generative models~\citep{kingma2013auto,rezende2014stochastic,ranganath2015deep,bamler2017dynamic}.

Before the advent of BBVI, divergence measures other than the KL divergence had been of limited practical use due to their complexity in both mathematical derivation and computation~\citep{minka2005divergence}, but have since then been revisited. Alpha-divergences~\citep{hernandez2016black,dieng2016chi,li2016renyi} achieve a better matching of the variational distribution to different regions of the posterior and may be tuned to either fit its dominant mode or to cover its entire support. The problem with reparameterizing the gradient of the alpha-divergence is, however, that the resulting gradient estimates have large variances. It is therefore desirable to find other divergence measures with low-variance reparameterization gradients.

In this paper, we use concepts from perturbation theory of statistical physics to propose a new family of variational bounds on the marginal likelihood with low-variance reparameterization gradients.
The lower bounds are enumerated by an order $K$, which takes odd integer values, and are given by
\begin{align}\label{eq:lower-bound-general-order}
	\mathcal L^{(K)}(\lambda, V_0)
    &= e^{-V_0} \sum_{k=0}^K \frac{1}{k!} \, \E_{\mathbf z\sim q}\!\left[\big( \!\log p(\bx, \bz) - \log q(\bz;\lambda) + V_0 \big)^k \right].
\end{align}
Here, $p(\bx,\bz)$ denotes the joint probability density function of the model with observations $\bx$ and latent variables $\bz$, $q$ is the variational distribution, which depends on variational parameters $\lambda$, and $V_0\in\mathbb R$ is a reference point for the perturbative expansion, see below.
In this paper, we motivate and discuss Eq.~\ref{eq:lower-bound-general-order} (Section~\ref{sec:method}), and we analyze the properties of the proposed bound experimentally (Section~\ref{sec:experiments}).
Our contributions are as follows.
\begin{itemize}
\item
We establish a view on black box variational inference with generalized divergences as a form of \textit{biased importance sampling} (Section~\ref{sec:biased_IS}).
The choice of divergence allows us to trade-off between a low-variance stochastic gradient and loose bound, and a tight variational bound with higher-variance Monte-Carlo gradients.
As we explain below, importance sampling and point estimation are at opposite ends of this spectrum.
\item
We combine these insights with ideas from perturbation theory of statistical physics to motivate the objective function in Eq.~\ref{eq:lower-bound-general-order} (Section~\ref{sec:PCVI}).
We show that, for all odd $K$, $\mathcal L^{(K)}(\lambda,V_0)$ is a nontrivial lower bound on the marginal likelihood $p(\bx)$.
Thus, we propose the perturbative black box variational inference algorithm (PBBVI), which maximizes $\mathcal L^{(K)}(\lambda,V_0)$ over $\lambda$ and $V_0$ with stochastic gradient descent (SGD).
For $K=1$, our algorithm is equivalent to standard BBVI with the KL-divergence (KLVI).
On the variance-bias spectrum, KLVI is on the side of large bias and low gradient variance.
Increasing $K$ to larger odd integers allows us to gradually trade in some increase in the gradient variance for some reduction of the bias.
\item
We evaluate our PBBVI algorithm experimentally for the lowest nonstandard order $K=3$ (Section~\ref{sec:experiments}).
Compared to KLVI ($K=1$), our algorithm fits variational distributions that cover more of the mass of the true posterior.
Compared to alpha-VI, our experiments confirm that PBBVI uses gradient estimates with lower variance, and converges faster.
\end{itemize}

\section{Related work}
\label{sec:related}
Our approach is related to BBVI, VI with generalized divergences, and variational perturbation theory. We thus briefly discuss related work in these three directions.

\paragraph {Black box variational inference (BBVI).}~ BBVI has already been addressed in the introduction~\citep{salimans2013fixed,kingma2013auto,rezende2014stochastic,ranganath2014black, ruiz2016generalized}; it enables variational inference for many models. Our work builds upon BBVI in that BBVI makes a large class of new divergence measures between the posterior and the approximating distribution tractable. Depending on the divergence measure, BBVI may suffer from high-variance stochastic gradients. This is a practical problem that we aim to improve in this paper.

\paragraph{Generalized divergences measures.}~ Our work connects to generalized information-theoretic divergences~\citep{shun2012differential}. \citet{minka2005divergence} introduced a broad class of divergences for variational inference, including alpha-divergences. Most of these divergences have been intractable in large-scale applications until the advent of BBVI.  In this context, alpha-divergences were first suggested by ~\citet{hernandez2016black} for local divergence minimization, and later for global minimization by~\citet{li2016renyi} and~\citet{dieng2016chi}. As we show in this paper, alpha-divergences have the disadvantage of inducing high-variance gradients, since the ratio between posterior and variational distribution enters the bound polynomially instead of logarithmically. In contrast, our approach leads to a more stable inference scheme in high dimensions.

\paragraph{Variational perturbation theory.}~
Perturbation theory refers to methods that aim to truncate a typically divergent power series to a convergent series.
In machine learning, these approaches have been addressed from an information-theoretic perspective by~\citet{tanaka1999theory,tanaka2000information}.
Thouless-Anderson-Palmer (TAP) equations \citep{thouless1977solution} are a form of second-order perturbation theory.
They were originally developed in statistical physics to include perturbative corrections to the mean-field solution of Ising models.
They have been adopted into Bayesian inference in~\citep{plefka1982convergence} and were advanced by many authors~\citep{kappen2001second,paquet2009perturbation,opper2013perturbative,opper2015expectation}.
In variational inference, perturbation theory yields extra terms to the mean-field variational objective which are difficult to calculate analytically.
This may be a reason why the methods discussed are not widely adopted by practitioners.
In this paper, we emphasize the ease of including perturbative corrections in a black box variational inference framework.
Furthermore, in contrast to earlier formulations, our approach yields a strict lower bound to the marginal likelihood which can be conveniently optimized.
Our approach is different from the traditional variational perturbation formulation~\citep{kleinert2009path}, which generally does not result in a bound.

\section{Method}
\label{sec:method}
In this section, we present our main contributions. We first present our
view of black box variational inference (BBVI) as a form of biased importance sampling in Section~\ref{sec:biased_IS}.
With this view, we bridge the gap between variational inference and importance sampling.
In Section~\ref{sec:PCVI}, we introduce our family of new variational bounds, and analyze their properties further in Section~\ref{sec:theory}.

\begin{figure}
    \centering
    \begin{minipage}[t]{0.3\textwidth}
        \centering
      \includegraphics[width=\columnwidth, height=0.8\columnwidth ]{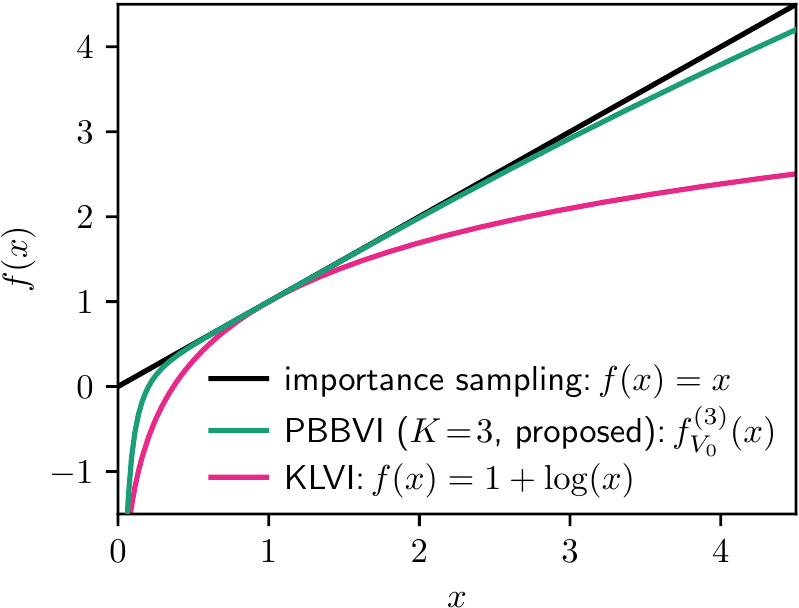}
        \captionof{figure}{\footnotesize Different choices for $f$ in Eq.~\ref{eq:biased-bound}. KLVI corresponds to $f(x)=\log(x) + \text{const.}$ (red), and importance sampling to $f(x) = x$ (black). Our proposed PBBVI bound uses $f_{V_0}^{(K)}$ (green, Eq.~\ref{eq:def_fc}), which lies between KLVI and importance sampling (we set $K=3$ and $V_0=0$ for PBBVI here).}
        \label{fig:f_lower_bound}
    \end{minipage}
    \hfill
    \begin{minipage}[t]{0.3\textwidth}
    \centering
    \includegraphics[width=\columnwidth, height=0.8\columnwidth ]{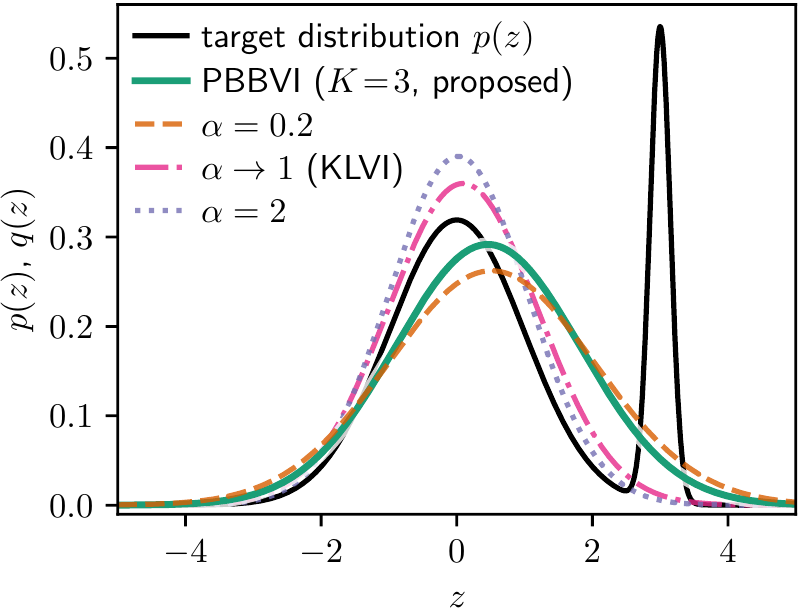}
    \captionof{figure}{\footnotesize Behavior of different VI methods on fitting a univariate Gaussian to a bimodal target distribution (black). PBBVI (proposed, green) covers more of the mass of the entire distribution than the traditional KLVI (red). Alpha-VI is mode seeking for large $\alpha$ and mass covering for smaller $\alpha$.}
    \label{fig:1DGaussian}
     \end{minipage}
     \hfill
    \begin{minipage}[t]{0.3\textwidth}
    \centering
    \includegraphics[width=\columnwidth, height=0.8\columnwidth ]{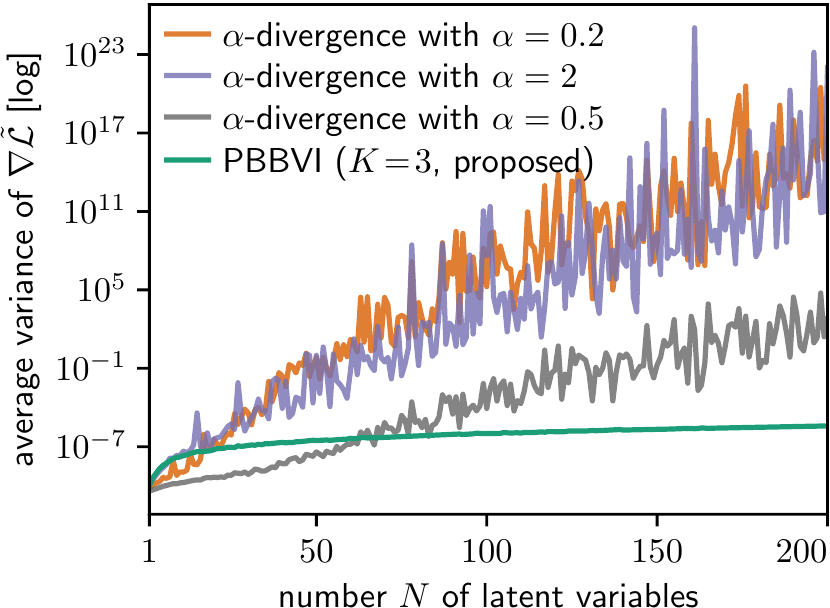}
    \captionof{figure}{\footnotesize Sampling variance of the stochastic gradient (averaged over its components) in the optimum, for alpha-divergences (orange, purple, gray), and the proposed PBBVI (green). The variance grows exponentially with the latent dimension $N$ for alpha-VI, and only algebraically for PBBVI.}
    \label{fig:variances}
     \end{minipage}
\end{figure}

\subsection{Black Box Variational Inference as Biased Importance Sampling}
\label{sec:biased_IS}

Consider a probabilistic model with data $\mathbf x$,  latent variables $\mathbf z$, and joint distribution $p(\mathbf x,\mathbf z)$.
We are interested in the posterior distribution over the latent variables, $p(\mathbf z | \mathbf x) = p(\mathbf x,\mathbf z) / p(\bx)$. This involves the intractable marginal likelihood $p(\bx)$. In variational inference~\citep{jordan1999introduction}, we instead minimize a divergence measure between a variational distribution $q(\bz;\lambda)$ and the posterior. Here, $\lambda$ are parameters of the variational distribution, and we aim to find the parameters $\lambda^*$ that minimize the distance to the posterior. This is equivalent to maximizing a lower bound on the marginal likelihood.

We call the difference between the log variational distribution and the log joint distribution the \textit{interaction energy},
\begin{align}
\label{eq:def_V}
V(\bz; \lambda) = \log q(\bz;\lambda) - \log p(\bx,\bz).
\end{align}
We use $V$ or $V(\bz)$ interchangeably to denote $V(\bz; \lambda)$, and $q(\bz)$ to denote $q(\bz;\lambda)$, when more convenient.
Using this notation, the marginal likelihood is
\begin{align}
\label{eq:IS_1}
p(\bx) =  \E_{q(\bz)}[e^{-V(\bz)}].
\end{align}
We call $e^{-V(\bz)} = p(\bx,\bz)/q(\bz)$  the \emph{importance ratio}, since
sampling from $q(\bz)$ to estimate the right-hand side of Eq.~\ref{eq:IS_1} is equivalent to importance sampling. As importance sampling is inefficient in high dimensions, we resort to variational inference.
To this end, let $f(\cdot)$ be any concave function defined on the positive reals. We assume furthermore that for all $x>0$, we have $f(x)\leq x$. Applying Jensen's inequality, we can lower-bound the marginal likelihood,
\begin{align}
\label{eq:biased-bound}
p(\bx) \geq f(p(\bx)) \geq  \E_{q(\bz)}[f(e^{-V(\bz;\lambda)})] \equiv {\cal L}_f(\lambda).
\end{align}

Figure \ref{fig:f_lower_bound} shows exemplary choices of $f$.
We maximize ${\cal L}_f(\lambda)$ using  reparameterization gradients, where the bound is not computed analytically, but rather its gradients are estimated by sampling from $q(\bz)$~\citep{kingma2013auto}. This leads to a stochastic gradient descent scheme, where the noise is a result of the Monte-Carlo estimation of the gradients.

Our approach builds on the insight that black box variational inference is a type of biased importance sampling, where we estimate a lower bound of the marginal likelihood by sampling from a proposal distribution, iteratively improving this distribution. The approach is biased since we do not estimate the exact marginal likelihood but only a lower bound to this quantity. As we argue below, the introduced bias allows us to estimate the bound more easily, because we decrease the variance of this estimator. The choice of the function $f$ thereby trades-off between bias and variance in the following way:
\begin{itemize}
\item For $f=id$ being the identity, we obtain \textit{importance sampling}. (See the black line in Figure~\ref{fig:f_lower_bound}). In this case, Eq.~\ref{eq:biased-bound} does not depend on the variational parameters, so there is nothing to optimize and we can directly sample from any proposal distribution $q$. Since the expectation under $q$ of the importance ratio $e^{-V(\bz)}$ gives the exact marginal likelihood, there is no bias. If the model has a large number of latent variables, the importance ratio $e^{-V(\bz)}$ becomes tightly peaked around the minimum of the interaction energy $V$, resulting in a very high variance of this estimator. Importance sampling is therefore on one extreme end of the bias-variance spectrum.
\item For $f = \log$, we obtain the familiar \emph{Kullback-Leibler (KL) bound}.
(See the pink line in Figure~\ref{fig:f_lower_bound}; here we add a constant of $1$ for comparison, which does not affect the optimization). Since $f(e^{-V(\bz)}) = -V(\bz)$, the bound is
\begin{align}\label{eq:klvi-elbo}
 {\cal L}_{KL}(\lambda) = \E_{q(\bz)}[-V(\bz)]=\E_{q(\bz)}[\log p(\bx,\bz) - \log q(\bz)].
\end{align}
The Monte-Carlo expectation of $\E_q[-V]$ has a much smaller variance than $\E_q[e^{-V}]$, implying efficient learning~\citep{bottou2010large}. However, by replacing $e^{-V}$ with $-V$ we introduce a bias. We can further trade-off less variance for even more bias by dropping the entropy term on the right-hand side of Eq.~\ref{eq:klvi-elbo}. A flexible enough variational distribution will shrink to zero variance, which completely eliminates the sampling noise. This is equivalent to point-estimation, and is at the opposite end of the bias-variance spectrum.
\item Now, consider any $f$ which is between the logarithm and the identity, e.g., the green line in Figure \ref{fig:f_lower_bound} (this is the regularizing function we propose in Section~\ref{sec:PCVI} below). The more similar $f$ is to the identity, the less biased is our estimate of the marginal likelihood, but the larger the variance. Conversely, the more $f$ behaves like the logarithm, the easier it is to estimate $f(e^{-V(\bz)})$ by sampling, while at the same time the bias grows.
\end{itemize}

One example of alternative divergences to the KL divergence that have been discussed in the literature are alpha-divergences~\citep{minka2005divergence, hernandez2016black,li2016renyi,dieng2016chi}.
Up to a constant, they correspond to the following choice of $f$:
\begin{align}
	f^{(\alpha)}(e^{-V}) &\propto  e^{-(1-\alpha) V}.
    \label{eq:alpha_divergences}
\end{align}
The real parameter $\alpha$ determines the distance to the importance sampling case ($\alpha=0$).
As $\alpha$ approaches $1$ from below, this bound leads to a better-behaved estimation problem of the Monte-Carlo gradient. However, unless taking the limit of $\alpha\rightarrow 1$ (where the objective becomes the KL-bound), $V$ still enters exponentially in the bound. As we show, this leads to a high variance of the gradient estimator in high dimensions (see Figure~\ref{fig:variances} discussed below). The alpha-divergence bound is therefore similarly as hard to estimate as the marginal likelihood in importance sampling.

Our analysis relies on the observation that expectations of exponentials in $V$ are difficult to estimate, and expectations of polynomials in $V$ are easy to estimate. We derive a family of new variational bounds which are polynomials in $V$, where increasing the order of the polynomial reduces the bias.

\subsection{Perturbative Black Box Variational Inference}
\label{sec:PCVI}
\paragraph{Perturbative bounds.}~
We now motivate the family of lower bounds proposed in Eq.~\ref{eq:lower-bound-general-order} in the introduction based on the considerations outlined above.
For fixed odd integer $K$ and fixed real value $V_0$, the bound $\mathcal L^{(K)}(\lambda,V_0)$ is of the form of Eq.~\ref{eq:biased-bound} with the following regularizing function $f$:
\begin{align} \label{eq:def_fc}
f_{V_0}^{(K)}(x) = e^{-V_0} \sum_{k=0}^K \frac{(V_0 + \log x)^k}{k!}
\qquad\Longrightarrow\qquad
f_{V_0}^{(K)}(e^{-V}) = e^{-V_0} \sum_{k=0}^K \frac{(V_0-V)^k}{k!}.
\end{align}
Here, the second (equivalent) formulation makes it explicit that $f_{V_0}^{(K)}$ is the $K$\textsuperscript{th} order Taylor expansion of its argument $e^{-V}$ in $V$ around some reference energy $V_0$.
Figure~\ref{fig:f_lower_bound} shows $f_{V_0}^{(K)}(x)$ for $K=1$ (red) and $K=3$ (green).
The curves are concave and lie below the identity, touching it at $x=e^{-V_0}$.
We show in Section~\ref{sec:theory} that these properties extend to every odd $K$ and every $V_0\in\mathbb R$.
Therefore, $\mathcal L^{(K)}(\lambda,V_0)$ is indeed a lower bound on the marginal likelihood.

The rationale for the design of the regularizing function in Eq.~\ref{eq:def_fc} is as follows.
On the one hand, the gradients of the resulting bound should be easy to estimate via the reparameterization approach.
We achieve low-variance gradient estimates by making $f_{V_0}^{(K)}(e^{-V})$ a polynomial in $V$, i.e., in contrast to the alpha-bound, $V$ never appears in the exponent.

On the other hand, the regularizing function should be close to the identity function so that the resulting bound has low bias.
For $K=1$, we have $\mathcal L^{(1)}(\lambda,V_0) = e^{-V_0}\E_q[\log p - \log q +V_0]$. Maximizing $\mathcal L^{(1)}$ over $\lambda$ is independent of the value of $V_0$ and equivalent to maximizing the standard KL bound $\mathcal L_{KL}$, see Eq.~\ref{eq:klvi-elbo}, which has low gradient variance and large bias.
Increasing the order $K$ to larger odd integers makes the Taylor expansion tighter, leading to a bound with lower bias.
In fact, in the limit $K\to\infty$, the right-hand side of Eq.~\ref{eq:def_fc} is the series representation of the exponential function, and thus $f_{V_0}^{(K)}$ converges pointwise to the identity.
In practice, we propose to set $K$ to a small odd integer larger than~$1$.
Increasing $K$ further reduces the bias, but it comes at the cost of increasing the gradient variance because the random variable $V$ appears in higher orders under the expectation in Eq.~\ref{eq:biased-bound}.

As discussed in Section~\ref{sec:biased_IS}, the KL bound $\mathcal L_{KL}$ can be derived from a regularizing function $f=\log$ that does not depend on any further parameters like $V_0$.
The derivation of the KL bound therefore does not require the first inequality in Eq.~\ref{eq:biased-bound}, and one directly obtains a bound on the model evidence $\log p(\bx) \equiv f(p(\bx))$ from the second inequality alone.
For $K>1$, the bound $\mathcal L^{(K)}(\lambda,V_0)$ depends nontrivially on $V_0$, and we have to employ the first inequality in Eq.~\ref{eq:biased-bound} in order to make the bounded quantity on the left-hand side independent of $V_0$.
This expenses some tightness of the bound but makes the method more flexible by allowing us to optimize over $V_0$ as well, as we describe next.

\begin{algorithm}[t]
\KwIn{joint probability $p(\bx,\bz)$;
 order of perturbation $K$ (odd integer);
 learning rate schedule $\rho_t$;
 number of Monte Carlo samples $S$;
 number of training iterations $T$;
 variational family $q(\bz,\lambda)$ that allows for reparameterization gradients, i.e., $\bz\sim q(\,\boldsymbol\cdot\,,\lambda) \Longleftrightarrow \bz = g(\boldsymbol\epsilon,\lambda)$ where $\boldsymbol\epsilon\sim p_\text{n}$ with a fixed noise distribution $p_\text{n}$ and a differentiable reparameterization function $g$.}\vspace{1pt}
\KwOut{fitted variational parameters $\lambda^*$.}\vspace{2pt}
\nl initialize $\lambda$ randomly and $V_0\gets0$\;
\nl \For{$t\gets1$ \KwTo $T$}{
\nl  draw $S$ samples $\boldsymbol\epsilon_1,\ldots,\boldsymbol\epsilon_S \sim p_\text{n}$ from the noise distribution\;\vspace{3pt}
  \textit{// obtain reparameterization gradient estimates using automatic differentiation:}\\
\nl  $g_\lambda \hspace{3.9pt}\!\gets \hat\nabla_{\!\lambda} \tilde{\mathcal L}^{(K)}(\lambda,V_0)
  \hspace{3.9pt}\equiv \nabla_{\!\lambda}\hspace{3.9pt} \!\Big[\! \frac{1}{S}\! \sum_{s=1}^S \sum_{k=0}^{K} \frac{1}{k!}\big(\!\log p(\bx,g(\boldsymbol\epsilon_s,\lambda)) - \log q(g(\boldsymbol\epsilon_s,\lambda);\lambda) + V_0 \big)^k\Big]$\;\vspace{-9pt}
\nl  $g_{V_0} \!\gets \hat\nabla_{\!V_0} \tilde{\mathcal L}^{(K)}(\lambda,V_0)
  \equiv \nabla_{\!V_0} \!\Big[\!\frac{1}{S}\! \sum_{s=1}^S \sum_{k=0}^{K} \frac{1}{k!} \big(\!\log p(\bx,g(\boldsymbol\epsilon_s,\lambda)) - \log q(g(\boldsymbol\epsilon_s,\lambda);\lambda) + V_0\big)^k$\Big]\;\vspace{-6pt}
  \textit{// perform variable updates (see second to last paragraph of Section~\ref{sec:PCVI}):}\\
\nl  $\lambda \hspace{4.4pt}\gets \lambda + \rho_t g_\lambda$\label{ln:update-lambda}\;
\nl  $V_0 \gets V_0 + \rho_t \Big[ g_{V_0} - \frac{1}{S} \sum_{s=1}^S \sum_{k=0}^{K} \frac{1}{k!} \big( \!\log p(\bx,g(\boldsymbol\epsilon_s,\lambda)) - \log q(g(\boldsymbol\epsilon_s,\lambda);\lambda) + V_0 \big)^k\Big]$\label{ln:update-v0}\;
 }
 \caption{Perturbative Black Box Variational Inference (PBBVI)}
 \label{alg:pbbvi}
\end{algorithm}

\paragraph{Optimization algorithm.}~
We now propose the perturbative black box variational inference (PBBVI) algorithm.
Since $\mathcal L^{(K)}(\lambda,V_0)$ is a lower bound on the marginal likelihood for all $\lambda$ and all $V_0$, we can find the values $\lambda^*$ and $V_0^*$ for which the bound is tightest by maximizing simultaneously over $\lambda$ and $V_0$.
Algorithm~\ref{alg:pbbvi} summarizes the PBBVI algorithm.
We minimize $-\mathcal L^{(K)}(\lambda,V_0)$ using stochastic gradient descent (SGD) with reparameterization gradients and a learning rate $\rho_t$ that decreases with the training iteration $t$ according to Robbins-Monro bounds~\citep{robbins1951stochastic}.
We obtain unbiased gradient estimators (denoted by ``$\hat\nabla$'') using standard techniques: we replace the expectation $\E_q[\,\cdot\,]$ in Eq.~\ref{eq:lower-bound-general-order} with the empirical average over a fixed number of $S$ samples from $q$, and we calculate the reparameterization gradients with respect to $\lambda$ and $V_0$ using automatic differentiation.

In practice, we typically discard the value of $V_0^*$ once the optimization is converged since we are only interested in the fitted variational parameters $\lambda^*$.
However, during the optimization process, $V_0$ is an important auxiliary quantity and the inference algorithm would be inconsistent without an optimization over $V_0$:
if we were to extend the model $p(\bx,\bz)$ by an additional observed variable $\tilde x$ which is statistically independent of the latent variables $\bz$, then the log joint (as a function of $\bz$ alone) changes by a constant positive prefactor.
The posterior remains unchanged by the constant prefactor, and a consistent VI algorithm must therefore produce the same approximate posterior distribution $q$ for both models.
Optimizing over $V_0$ ensures this consistency since the log joint appears in the lower bound only in the combination $\log p(\bx,\bz) + V_0$.
Therefore, a rescaling of the log joint by a constant positive prefactor can be completely absorbed by a change in the reference energy $V_0$.

We observed in our experiments that the reference energy $V_0$ can become very large (in absolute value) for models with many latent variables.
To avoid numerical overflow or underflow from the prefactor $e^{-V_0}$, we consider the surrogate objective $\tilde {\cal L}^{(K)}(\lambda,V_0) \equiv e^{V_0}{\cal L}^{(K)}(\lambda,V_0)$.
The gradients with respect to $\lambda$ of ${\cal L}^{(K)}(\lambda,V_0)$ and $\tilde {\cal L}^{(K)}(\lambda,V_0)$ are equal up to a positive prefactor, so we can replace the former with the latter in the update step (line~\ref{ln:update-lambda} in Algorithm~\ref{alg:pbbvi}).
The gradient with respect to $V_0$ is $\nabla_{\!V_0} {\cal L}^{(K)}(\lambda,V_0) \propto \nabla_{\!V_0} \tilde {\cal L}^{(K)}(\lambda,V_0) -\tilde {\cal L}^{(K)}(\lambda,V_0)$ (line~\ref{ln:update-v0}).
Using the surrogate $\tilde{\cal L}^{(K)}(\lambda,V_0)$ avoids numerical underflow or overflow, as well as exponentially increasing or decreasing gradients.

\paragraph{Mass covering effect.}~
In Figure~\ref{fig:1DGaussian}, we fit a Gaussian distribution to a one-dimensional bimodal target distribution (black line), using different divergences. Compared to BBVI with the standard KL divergence (KLVI, pink line), alpha-divergences are more mode-seeking (purple line) for large values of $\alpha$, and more mass-covering (orange line) for small $\alpha$~\citep{li2016renyi}. Our PBBVI bound ($K=3$, green line) achieves a similar mass-covering effect as in alpha-divergences, but with associated low-variance reparameterization gradients. This is seen in Figure~\ref{fig:variances}, discussed in Section~\ref{sec:GP_classification}, which compares the gradient variances of alpha-VI and PBBVI as a function of dimensions.

\subsection{Proof of Correctness and Nontriviality of the Bound}
\label{sec:theory}

To conclude the presentation of the PBBVI algorithm, we prove that the objective in Eq.~\ref{eq:lower-bound-general-order} is indeed a lower bound on the marginal likelihood for all odd orders $K$, and that the bound is nontrivial.

\paragraph{Correctness.}~
The lower bound $\mathcal L^{(K)}(\lambda,V_0)$ results from inserting the regularizing function $f_{V_0}^{(K)}$ from Eq.~\ref{eq:def_fc} into Eq.~\ref{eq:biased-bound}.
For odd $K$, it is indeed a valid lower bound because $f_{V_0}^{(K)}$ is concave and lies below the identity.
To see this, note that the second derivative $\partial^2 f_{V_0}^{(K)}(x)/\partial x^2= -e^{-V_0}(V_0+\log x)^{K-1}/((K-1)!\, x^2)$ is non-positive everywhere for odd $K$.
Therefore, the function is concave.
Next, consider the function $g(x)=f_{V_0}^{(K)}(x) - x$, which has a stationary point at $x=x_0\equiv e^{-V_0}$.
Since $g$ is also concave, $x_0$ is a global maximum, and thus $g(x) \leq g(x_0)=0$ for all $x$, implying that $f_{V_0}^{(K)}(x)\leq x$.
Thus, for odd $K$, the function $f_{V_0}^{(K)}$ satisfies all requirements for Eq.~\ref{eq:biased-bound}, and ${\cal L}^{(K)}(\lambda,V_0)\equiv\E_q[f_{V_0}^{(K)}(e^{-V})]$ is a lower bound on the marginal likelihood.
Note that an even order $K$ does not lead to a valid concave regularizing function.

\paragraph{Nontriviality.}~
Since the marginal likelihood $p(\bx)$ is always positive, a lower bound would be trivial if it was negative.
We show that once the optimization algorithm has converged, the bound at the optimum is always positive.
At the optimum, all gradients vanish.
By setting the derivative with respect to $V_0$ of the right-hand side of Eq.~\ref{eq:lower-bound-general-order} to zero we find that $\E_{q^*}[(V_0^*-V)^K]=0$, where $q^*\equiv q(\,\boldsymbol\cdot\,;\lambda^*)$ is the variational distribution at the optimum.
Thus, the lower bound at the optimum is $\mathcal L(\lambda^*,V_0^*) = e^{-V_0} \E_{q^*}\!\left[ h(V) \right]$ with $h(V) = \sum_{k=0}^{K-1} \frac{1}{k!}( V_0^* - V )^k$, where the sum runs only to $K-1$ because the term with $k=K$ vanishes at $V_0=V_0^*$.
We show that $h(V)$ is positive for all $V$.
If $K=1$, then $h(V)=1$ is a positive constant.
For $K\geq3$, $h(V)$ is a polynomial in $V$ of even order $K-1$, whose highest order term has a positive coefficient $1/(K-1)!$.
Therefore, as $V\to\pm\infty$, the function $h(V)$ goes to positive infinity and it thus has a global minimum at some value $\tilde V\in\mathbb R$.
At the global minimum, its derivative vanishes, $0 = \nabla_{\tilde V} h(\tilde V) = -\sum_{k=0}^{K-2} \frac{1}{k!}( V_0^* - \tilde V )^k$.
Thus, at the global minimum of the polynomial $h$, all terms except the highest order term cancel, and we find $h(\tilde V) = \frac{1}{(K-1)!}(V_0^* - \tilde V)^{K-1}\geq0$, which is nonnegative because $K-1$ is even.
The case $h(\tilde V)=0$ is achieved if and only if $\tilde V= V_0^*$, but this would violate the condition $\nabla_{\tilde V} h(\tilde V)=0$.
Therefore, $h(\tilde V)$ is strictly positive, and since $\tilde V$ is a global minimum of $h$, we have $h(V) \geq h(\tilde V) > 0$ for all $V\in\mathbb R$.
Inserting into the expression for $\mathcal L(\lambda^*,V_0^*)$ concludes the proof that the lower bound at the optimum is positive.

\section{Experiments}
\label{sec:experiments}
We evaluate PBBVI with different models. First we investigate its behavior in a controlled setup of Gaussian processes on synthetic data (Section \ref{sec:synthetic}). We then evaluate PBBVI based on a classification task using Gaussian processes classifiers, where we use data from the UCI machine learning repository (Section \ref{sec:GP_classification}). This is a Bayesian non-conjugate setup where black box inference is required. Finally, we use an experiment with the variational autoencoder (VAE) to explore our approach on a deep generative model (Section \ref{sec:VAE}). This experiment is carried out on  MNIST data.
We use the perturbative order $K=3$ for all experiments with PBBVI.
This corresponds to the lowest order beyond standard KLVI, since KLVI is equivalent to PBBVI with $K=1$, and $K$ has to be an odd integer.
Across all the experiments, PBBVI demonstrates advantages based on different metrics.

\subsection{GP Regression on Synthetic Data}
\label{sec:synthetic}
\begin{figure}[t]
\centering
\subfigure[KLVI]{\includegraphics[width=0.42 \textwidth]{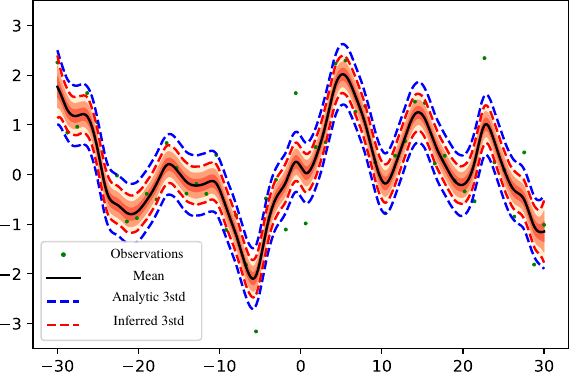}}
\hspace{15pt}
\subfigure[PBBVI with $K=3$]{\includegraphics[width=0.42 \textwidth]{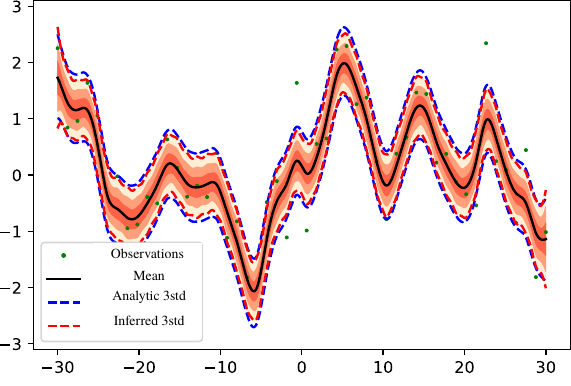}}
\caption{ \footnotesize Gaussian process regression on synthetic data (green dots). Three standard deviations are shown in varying shades of orange.
The blue dashed lines show three standard deviations of the true posterior. The red dashed lines show the inferred three standard deviations using KLVI (a) and PBBVI (b). We see that the results from our proposed PBBVI are close to the analytic solution while traditional KLVI underestimates the variances.}
\label{fig:Toy_GP_fig}
\vspace{4pt}
\end{figure}

\begin{table}[t]
\parbox{.43\linewidth}{
\centering
\begin{tabular}{l c }
 \toprule
 Method &Avg variances\\
 \cmidrule(lr){1-2}
 Analytic & 0.0415 \\
 KLVI & 0.0176 \\
 PBBVI &  0.0355 \\
 \bottomrule
\end{tabular}
\medskip
\caption{ \footnotesize Average variances across training examples in the synthetic data experiment. The closer to the analytic solution, the better.}
\label{tab:Toy_GP_tab}
}
\hfill
\parbox{.53\linewidth}{
\centering
\begin{tabular}{ l  c c c c}
&&&&\\
 \toprule
Data set &Crab &Pima & Heart & Sonar   \\
 \cmidrule(lr){1-5}
KLVI &0.22& 0.245   & 0.148 & 0.212 \\
PBBVI & \textbf{0.11} & \textbf{0.240}  & \textbf{0.1333} & \textbf{0.1731} \\
 \bottomrule
\end{tabular}
\medskip
\caption{\footnotesize Error rate of GP classification on the test set. The lower the better. Our proposed PBBVI consistently obtains better classification results. }
\label{tab:UCI_tab}
}
\vspace{-14pt}
\end{table}

In this section, we inspect the inference behavior using a synthetic data set with Gaussian processes (GP).  We generate the data according to a Gaussian noise distribution centered around a mixture of sinusoids, and sample 50 data points (green dots in Figure~\ref{fig:Toy_GP_fig}).
We then use a GP to model the data, thus assuming the generative process $f \sim \mathcal{GP} (0, \Lambda)$ and $y_i \sim \mathcal{N}(f_i, \epsilon)$.

We first compute an analytic solution of the posterior of the GP, (three standard deviations shown in blue dashed lines) and compare it to approximate posteriors obtained by KLVI (Figure \ref{fig:Toy_GP_fig} (a)) and  the proposed PBBVI (Figure \ref{fig:Toy_GP_fig} (b)). The results from PBBVI are almost identical to the analytic solution. In contrast, KLVI underestimates the posterior variance. This is consistent with Table \ref{tab:Toy_GP_tab}, which shows the average diagonal variances. PBBVI results are much closer to the exact posterior variances.

\subsection{Gaussian Process Classification}
\label{sec:GP_classification}

We evaluate the performance of PBBVI and KLVI on a GP classification task. Since the model is non-conjugate, no analytical baseline is available in this case. We model the data with the following generative process:
\begin{equation*}
f \sim \mathcal{GP}(0, \Lambda),~~~~ z_i = \sigma(f_i),~~~~ y_i \sim Bern(z_i).
\end{equation*}
Above, $\Lambda$ is the GP kernel, $\sigma$ indicates the sigmoid function, and $Bern$ indicates the Bernoulli distribution. We furthermore use the Matern 32 kernel,
\begin{equation*}
\Lambda_{ij} = s^2 (1 + \textstyle \frac{\sqrt{3}\, r_{ij}}{l}) \exp( - \textstyle \frac{\sqrt{3}\, r_{ij}}{l}),~~~~
r_{ij}=\sqrt{(x_i-x_j)^T(x_i - x_j)}.
\end{equation*}

\paragraph{Data.}~ We use four data sets from the UCI machine learning repository, suitable for binary classification: Crab (200 datapoints), Pima (768 datapoints), Heart (270 datapoints), and Sonar (208 datapoints).
We randomly split each of the data sets into two halves. One half is used for training and the other half is used for testing. We set the hyper parameters $s = 1$ and $l = \sqrt{D}/2$ throughout all experiments, where $D$ is the dimensionality of input $x$.

Table \ref{tab:UCI_tab} shows the classification performance (error rate) for these data sets. Our proposed PBBVI consistently performs better than the traditional KLVI.

\paragraph{Convergence speed comparison.}~ We also carry out a comparison in terms of speed of convergence, focusing on PBBVI and alpha-divergence VI.
Our results indicate that the smaller variance of the reparameterization gradient leads to faster convergence of the optimization algorithm.

We train the GP classifier from Section~\ref{sec:GP_classification} on the Sonar UCI data set using a constant learning rate.
Figure \ref{fig:convergence} shows the test log-likelihood under the posterior mean as a function of training iterations.
We split the data set into equally sized training, validation, and test sets.
We then tune the learning rate and the number of Monte Carlo samples per gradient step to obtain optimal performance on the validation set after minimizing the alpha-divergence with a fixed budget of random samples.
We use $\alpha=0.5$ here; smaller values of $\alpha$ lead to even slower convergence.
We optimize the PBBVI lower bound using the same learning rate and number of Monte Carlo samples.
The final test error rate is $22\%$ on an approximately balanced data set.
PBBVI converges an order of magnitude faster.

Figure~\ref{fig:variances} in Section~\ref{sec:method} provides more insight in the scaling of the gradient variance.
Here, we fit GP regression models on synthetically generated data by maximizing the PBBVI lower bound and the alpha-VI lower bound with $\alpha\in\{0.2, 0.5, 2\}$.
We generate a separate synthetic data set for each $N\in\{1,\ldots,200\}$ by drawing $N$ random data points around a sinusoidal curve.
For each $N$, we fit a one-dimensional GP regression with PBBVI and alpha-VI, respectively, using the same data set for both methods.
The variational distribution is a fully factorized Gaussian with $N$ latent variables.
After convergence, we estimate the sampling variance of the gradient of each lower bound with respect to the posterior mean.
We calculate the empirical variance of the gradient based on $10^5$ samples from $q$, and we average over the $N$ coordinates.
Figure~\ref{fig:variances} shows the average sampling variance as a function of $N$ on a logarithmic scale.
The variance of the gradient of the alpha-VI bound grows exponentially in the number of latent variables.
By contrast, we find only algebraic growth for PBBVI.

\subsection{Variational Autoencoder}
\label{sec:VAE}
\begin{figure}
    \centering
    \begin{minipage}{0.49\textwidth}
        \centering
      \includegraphics[height=0.5\columnwidth]{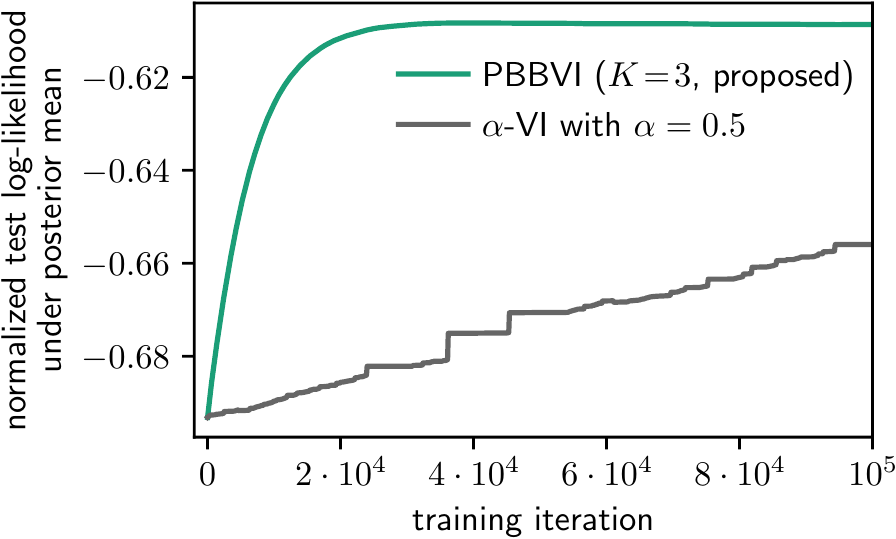}
        \caption{\footnotesize Test log-likelihood (normalized by the number of test points) as a function of training iterations using GP classification on the Sonar data set. PBBVI converges faster than alpha-VI even though we tuned the number of Monte Carlo samples per training step ($100$) and the constant learning rate ($10^{-5}$) so as to maximize the performance of alpha-VI on a validation set.}
        \label{fig:convergence}
    \end{minipage}
     \hfill
    \begin{minipage}{0.49\textwidth}
    \centering
    \includegraphics[height=0.5\columnwidth]{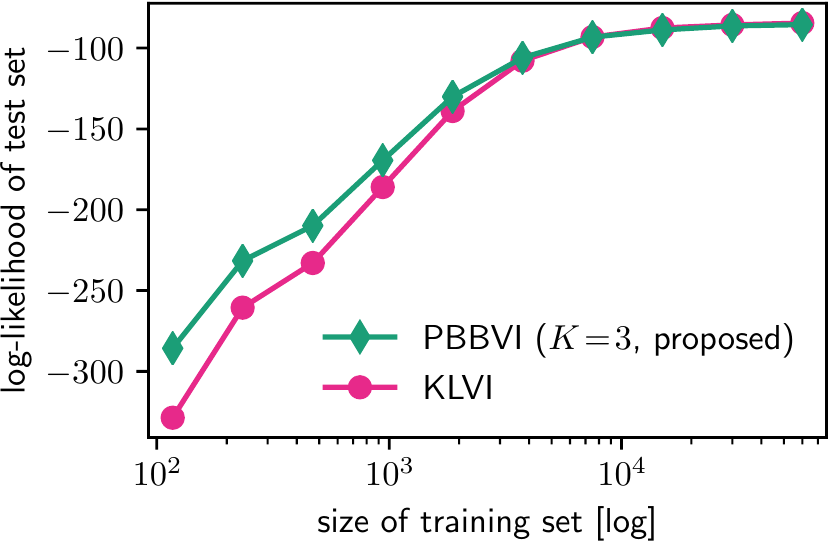}
    \caption{\footnotesize Predictive likelihood of a VAE trained on different sizes of the data. The training data are randomly sampled subsets of the MNIST training set. The higher value the better. Our proposed PBBVI method outperforms KLVI mainly when the size of the training data set is small. The fewer the training data, the more advantage PBBVI obtains.}
    \label{fig:MNIST}
     \end{minipage}
\end{figure}

We experiment on Variational Autoencoders (VAEs), and we compare the PBBVI and the KLVI bound in terms of predictive likelihoods on held-out data~\citep{kingma2013auto}.
Autoencoders compress unlabeled training data into low-dimensional representations by fitting it to an encoder-decoder model that maps the data to itself.
These models are prone to learning the identity function when the hyperparameters are not carefully tuned, or when the network is too expressive, especially for a moderately sized training set.
VAEs are designed to partially avoid this problem by estimating the uncertainty that is associated with each data point in the latent space.
It is therefore important that the inference method does not underestimate posterior variances.
We show that, for small data sets, training a VAE by maximizing the PBBVI lower bound leads to higher predictive likelihoods than maximizing the KLVI lower bound.

We train the VAE on the MNIST data set of handwritten digits~\citep{lecun1998gradient}.
We build on the publicly available implementation by~\citet{burda2015importance} and also use the same architecture and hyperparamters, with $L=2$ stochastic layers and $S=5$ samples from the variational distribution per gradient step.
The model has $100$ latent units in the first stochastic layer and $50$ latent units in the second stochastic layer.

The VAE model factorizes over all data points.
We train it by stochastically maximizing the sum of the PBBVI lower bounds for all data points using a minibatch size of $20$.
The VAE amortizes the gradient signal across data points by training inference networks.
The inference networks express the mean and variance of the variational distribution as a function of the data point.
We add an additional inference network that learns the mapping from a data point to the reference energy $V_0$.
Here, we use a network with four fully connected hidden layers of $200$, $200$, $100$, and $50$ units, respectively.

MNIST contains $60{,}000$ training images.
To test our approach on smaller-scale data where Bayesian uncertainty matters more, we evaluate the test likelihood after training the model on randomly sampled fractions of the training set.
We use the same training schedules as in the publicly available implementation, keeping the total number of training iterations independent of the size of the training set.
Different to the original implementation, we shuffle the training set before each training epoch as this turns out to increase the performance for both our method and the baseline.

Figure \ref{fig:MNIST} shows the predictive log-likelihood of the whole test set, where the VAE is trained on random subsets of different sizes of the training set.
We use the same subset to train with PBBVI and KLVI for each training set size.
PBBVI leads to a higher predictive likelihood than traditional KLVI on subsets of the data. We explain this finding with our observation that the variational distributions obtained from PBBVI capture more of the posterior variance.
As the size of the training set grows---and the posterior uncertainty decreases---the performance of KLVI catches up with PBBVI.

As a potential explanation why PBBVI converges to the KLVI result for large training sets, we note that $\E_{q^*}[(V_0^*-V)^3]=0$ at the optimal variational distribution $q^*$ and reference energy $V_0^*$ (see Section~\ref{sec:theory}).
If $V$ becomes a symmetric random variable (such as a Gaussian) in the limit of a large training set, then this implies that $\E_{q^*}[V]=V_0^*$, and PBBVI reduces to KLVI for large training sets.

\section{Conclusion}
\label{sec:conclusion}
We first presented a view on  black box variational inference as a form of biased importance sampling, where we can trade-off bias versus variance by the choice of divergence.
Bias refers to the deviation of the bound from the true marginal likelihood, and variance refers to its reparameterization gradient estimator.
We then proposed a family of new variational bounds that connect to variational perturbation theory, and which include corrections to the standard Kullback-Leibler bound.
Our proposed PBBVI bound converges to the true marginal likelihood for large order $K$ of the perturbative expansion, and we showed both theoretically and experimentally that it has lower-variance reparameterization gradients compared to alpha-VI.
In order to scale up our method to massive data sets, future work will explore stochastic versions of PBBVI.
Since the PBBVI bound contains interaction terms between all data points, breaking it up into mini-batches is non-straightforward.
Besides, while our experiments used a fixed perturbative order of $K=3$, it could be beneficial to increase the perturbative order at some point during the training cycle once an empirical estimate of the gradient variance drops below a certain threshold.
Furthermore, the PBBVI and alpha-bounds can also be combined, such that PBBVI further approximates alpha-VI.
This could lead to promising results on large data sets where traditional alpha-VI is hard to optimize due to its variance, and traditional PBBVI converges to KLVI.
As a final remark, a tighter variational bound is not guaranteed to always result in a better posterior approximation since the variational family limits the quality of the solution.
However, in the context of variational EM, where one performs gradient-based hyperparameter optimization on the log marginal likelihood, our bound gives more reliable results since higher orders of $K$ can be assumed to approximate the marginal likelihood better.
\clearpage
\bibliographystyle{apa}
\bibliography{ref}
\clearpage
\end{document}

\maketitle

\section*{Proof that PBBVI minimizes a divergence}

In this supplement we show that perturbative black box variational inference (PBBVI) minimizes a valid divergence from the variational distribution $q(\mathbf z)$ to the true posterior distribution $p(\mathbf z|\mathbf x)$.
This has the important consequence that PBBVI converges to exact inference in the limit of an arbitrarily flexible variational family.
In contrast to traditional Kullback-Leibler variational inference (KLVI) and $\alpha$-VI, the divergence minimized by PBBVI depends on the choice of variational family.

Let $f$ be any regularizing function that satisfies the conditions for Eq.~4 of the main text (i.e., $f$ is concave and smaller than the identity).
Let $\mathcal L_f(q)$ be the associated lower bound on the marginal likelihood.
We prefer this notation over the notation $\mathcal L_f(\lambda)$ used in the main text here, because, at this stage, we do not restrict $q$ to a specific family of variational distributions indexed by $\lambda$.
We define a divergence $D_f$ from $q$ to the true posterior $p(\mathbf z|\mathbf x)$ by
\begin{align}
  D_f(p || q) \equiv f(p(\mathbf x)) - \mathcal L_f(q).
\end{align}
Here, the model marginal likelihood $p(\mathbf x)$ is unknown to us, but it is a well-defined constant assuming that the model parameters are kept constant.
We show that $D_f$ is indeed a valid divergence.
From Eq.~4 of the main text, we find $\mathcal L_f(q) \leq f(p(\mathbf x))$ and therefore $D_f$ is non-negative.
The lower bound $\mathcal L_f(q)$ is defined in Eqs.~2 and 4 of our paper as
\begin{align}
  \mathcal L_f(q) \equiv \mathbb E_{q(\mathbf z)}\left[ f\!\left( \frac{p(\mathbf x, \mathbf z)}{q(\mathbf z)} \right) \right].
\end{align}
Setting $q(\mathbf z)$ to the true posterior, $p(\mathbf x,\mathbf z) / p(\mathbf x)$, yields $\mathcal L_f(q) = f(p(\mathbf x))$, and therefore sets $D_f(p || q)$ to zero.
Thus, $D_f$ is indeed a valid divergence.

Consider now the specific family of regularizing functions $f^{(K)}_{V_0}$ defined in Eq.~7 of the main text.
We now also restrict $q$ to be a member of some predefined variational family parameterized by $\lambda$.
Maximizing the corresponding lower bound $\mathcal L^{(K)}(\lambda,V_0) \equiv \mathcal L_{f_{V_0}^{(K)}}(q)$ simultaneously over $\lambda$ and $V_0$ yields an optimal reference energy $V_0^*$ and an optimal member $q^*\equiv q(\,\boldsymbol\cdot\,,\lambda^*)$ of the variational family.
Both depend not only on the model but also on the variational family to which $q$ is restricted.
Evidently, $q^*$ is the member of the variational family that minimizes the divergence
\begin{align}
  D_{f_{V_0^*}^{(K)}}(p || q) = f_{V_0^*}^{(K)}(p(\mathbf x)) - \mathcal L_{f_{V_0^*}^{(K)}}(q).
\end{align}
Here, the first term on the right-hand side is a constant (since $V_0^*$ is).
Its value is not known to us, but well defined.
Thus, PBBVI minimizes a valid divergence to the true posterior.
As a practical consequence this implies that the exact maximum of the PBBVI lower bound is the true posterior if the variational family is sufficiently flexible to contain it.
Note that the choice of divergence that PBBVI minimizes depends on the perturbative order $K$, and also on the model and the variational family (via their influence on $V_0^*$).